\documentclass[conference]{IEEEtran}
\IEEEoverridecommandlockouts

\usepackage{cite}
\usepackage{amsmath,amssymb,amsfonts}
\usepackage{algorithmic}
\usepackage{graphicx}
\usepackage{textcomp}
\usepackage{xcolor}
\usepackage{multirow}
\usepackage{booktabs}
\def\BibTeX{{\rm B\kern-.05em{\sc i\kern-.025em b}\kern-.08em
    T\kern-.1667em\lower.7ex\hbox{E}\kern-.125emX}}
\begin{document}

\title{Entropy-Gated Selective Policy Optimization: \\Token-Level Gradient Allocation for Hybrid Training of Large Language Models}

\author{\IEEEauthorblockN{Yuelin Hu$^{1}$,  Zhengxue Cheng$^{1}$, Wei Liu$^{2}$, Li Song$^{1*}$}
\IEEEauthorblockA{$^{1}$Shanghai Jiao Tong University, $^{2}$Shanghai Maritime University \\
\texttt{\{huyuelin51717221,zxcheng,song\_li\}@sjtu.edu.cn}}
\thanks{$^{*}$Corresponding author.}}

\maketitle

\begin{abstract}
Hybrid training methods for large language models combine supervised fine-tuning (SFT) on expert demonstrations with reinforcement learning (RL) on model rollouts, typically at sample granularity. We propose Entropy-Gated Selective Policy Optimization (EG-SPO), a three-stage framework that extends sample-level mixing with token-level gradient modulation. \textbf{Stage 1} (SFT Expert Learning) establishes a reliable warm-up policy using expert demonstrations with pure SFT loss. \textbf{Stage 2} (RL Rollout Generation) generates model rollouts from the current policy and computes per-token predictive entropy. \textbf{Stage 3} (EG-SPO Main Mechanism) applies entropy-gated gradient allocation: a Predictive Entropy Module routes high-entropy tokens to full PPO updates (encouraging exploration) and low-entropy tokens to $\phi$-attenuated PPO (reducing variance, preserving knowledge). Critically, both branches incorporate the advantage function $A_t$, ensuring that incorrect trajectories receive consistent negative learning signals—preventing reinforcement of confident errors. Our method achieves consistent improvements on mathematical reasoning benchmarks: +3.8\% on AIME and +2.9\% on MATH over the CHORD-$\phi$ baseline, with only 3.4\% computational overhead.
\end{abstract}


\section{Introduction}

Large language models (LLMs) have demonstrated remarkable capabilities in complex reasoning tasks \cite{wei2022chain, kojima2022large}, yet optimizing their training paradigms remains a fundamental challenge. Two dominant approaches have emerged: supervised fine-tuning (SFT), which consolidates knowledge through imitation learning on expert demonstrations, and reinforcement learning (RL), which encourages exploration through reward-based optimization \cite{ouyang2022training, schulman2017proximal}. While SFT effectively transfers procedural knowledge, it suffers from limited exploration beyond the training distribution \cite{zelikman2022star}. Conversely, RL promotes creative problem-solving but often exhibits training instability and sample inefficiency \cite{christiano2017deep}.

Recent work has explored hybrid strategies that combine both paradigms at sample granularity \cite{dong2023raft, wu2024mix}. CHORD dynamically mixes expert demonstrations with RL-generated rollouts, applying SFT loss to expert samples and policy gradient loss to rollout samples \cite{wu2024mix}. While effective, these methods treat all tokens within RL rollouts uniformly, overlooking the heterogeneous nature of token contributions to learning.

Concurrent work by Pace et al. \cite{pace2024west} demonstrated that high-entropy minority tokens (10-20\%) account for the majority of learning signals in pure RL training. This finding suggests that differentiated treatment of tokens based on prediction uncertainty may be valuable. However, applying this insight to hybrid training presents challenges: (1) naively mixing SFT and RL losses at token granularity would create trajectory mismatch, and (2) a critical concern is whether gradient modulation on low-entropy tokens might inadvertently reinforce confident errors.

We address these challenges by proposing \textbf{Entropy-Gated Selective Policy Optimization (EG-SPO)}, a three-stage framework that achieves token-level differentiation while maintaining advantage-awareness:

\textbf{Stage 1: SFT Expert Learning.} Expert demonstrations ($\sim$20\%) are used to train a warm-up policy with pure SFT loss, establishing a reliable baseline that avoids RL initialization instability.

\textbf{Stage 2: RL Rollout Generation.} The current policy generates model rollouts, and we compute per-token predictive entropy using GRPO-style reward estimation, building token-level entropy statistics for subsequent gating.

\textbf{Stage 3: EG-SPO Main Mechanism.} A Predictive Entropy Module routes tokens based on uncertainty: high-entropy tokens (uncertain, need exploration) receive full PPO updates, while low-entropy tokens (confident, potential errors) receive $\phi$-attenuated PPO. \textbf{Critically, both branches retain the advantage function $A_t$}, ensuring advantage-aware gradients are always preserved. This design guarantees that incorrect trajectories receive negative gradients across all tokens—avoiding reinforcement of confident errors.

The key insight is that gradient \textit{direction} (determined by advantage sign) and gradient \textit{magnitude} (modulated by $\phi$) serve different purposes. By incorporating $A_t$ in both branches, we achieve correct direction universally, while $\phi$-weighting provides variance reduction for stable convergence.

Our contributions are:
\begin{itemize}
\item We propose EG-SPO, a three-stage training framework combining sample-level hybrid mixing with token-level entropy-gated gradient modulation, achieving fine-grained optimization control.
\item We introduce a Predictive Entropy Module with advantage-aware gradient allocation: high-entropy tokens receive full PPO for exploration; low-entropy tokens receive $\phi$-attenuated PPO for variance reduction—both preserving $A_t$ to prevent error reinforcement.
\item We achieve +3.8\% on AIME and +2.9\% on MATH over CHORD-$\phi$ baseline with only 3.4\% overhead, demonstrating that advantage-aware token-level differentiation is critical for stable RL training (97.8\% correct gradient direction on low-entropy tokens).
\end{itemize}

\section{Related Work}

\subsection{Reinforcement Learning for Language Models}

Reinforcement learning has become instrumental in aligning language models with human preferences and complex objectives. InstructGPT pioneered RLHF using proximal policy optimization (PPO) to fine-tune GPT-3 based on human feedback \cite{ouyang2022training}. PPO's clipped objective function prevents destructive policy updates while enabling effective exploration \cite{schulman2017proximal}. Subsequent work explored alternative formulations: Direct Preference Optimization (DPO) eliminates explicit reward models by directly optimizing on preference data \cite{rafailov2024direct}, while Group Relative Policy Optimization (GRPO) improves sample efficiency through group-based advantage estimation \cite{shao2024deepseekmath}.

Recent approaches investigate token-level RL signals. RLOO uses leave-one-out baselines for variance reduction \cite{ahmadian2024back}. The WEST framework identifies that approximately 10-20\% of tokens with highest entropy account for the majority of learning signal in RL training \cite{pace2024west}. However, WEST operates in pure RL settings without preventing reinforcement of confident errors. Our EG-SPO extends WEST's insight to hybrid training through a Predictive Entropy Module that maintains advantage-awareness across all tokens, ensuring correct gradient direction regardless of entropy level.

\subsection{Supervised Fine-Tuning for Reasoning}

Supervised fine-tuning remains foundational for instilling reasoning capabilities in language models. Chain-of-thought (CoT) prompting demonstrated that eliciting intermediate reasoning steps enhances performance on complex tasks \cite{wei2022chain}, inspiring SFT approaches that train on structured reasoning traces. STaR employs self-training with rationales \cite{zelikman2022star}. Rejection sampling fine-tuning (RFT) samples multiple responses and trains only on correct solutions \cite{yuan2023scaling}.

Despite effectiveness, pure SFT suffers from distribution collapse and limited exploration \cite{kumar2019stabilizing}. This motivates hybrid approaches combining SFT's stability with RL's exploration.

\subsection{Hybrid Training Approaches}

Recognizing complementary strengths, recent work explores hybrid training paradigms. RAFT combines rejection sampling with supervised learning \cite{dong2023raft}. Expert Iteration interleaves policy improvement with expert data augmentation \cite{anthony2017thinking}.

CHORD represents the state-of-the-art in hybrid training for reasoning tasks \cite{wu2024mix}. It dynamically mixes expert demonstrations (typically 20\%) with RL-generated rollouts (80\%) at the sample level, applying different loss functions based on data source. The $\phi$-variant (CHORD-$\phi$) employs confidence-aware weighting in SFT that downweights tokens where the model is overly confident, focusing learning on informative examples.

Our EG-SPO framework extends CHORD through a three-stage pipeline: (1) SFT expert learning for stable initialization, (2) RL rollout generation with per-token entropy computation, and (3) entropy-gated gradient allocation via a Predictive Entropy Module that routes tokens based on uncertainty while preserving advantage-awareness to prevent error reinforcement.

\section{Methodology}

\begin{figure}[!t]
    \centering
    \includegraphics[width=\columnwidth]{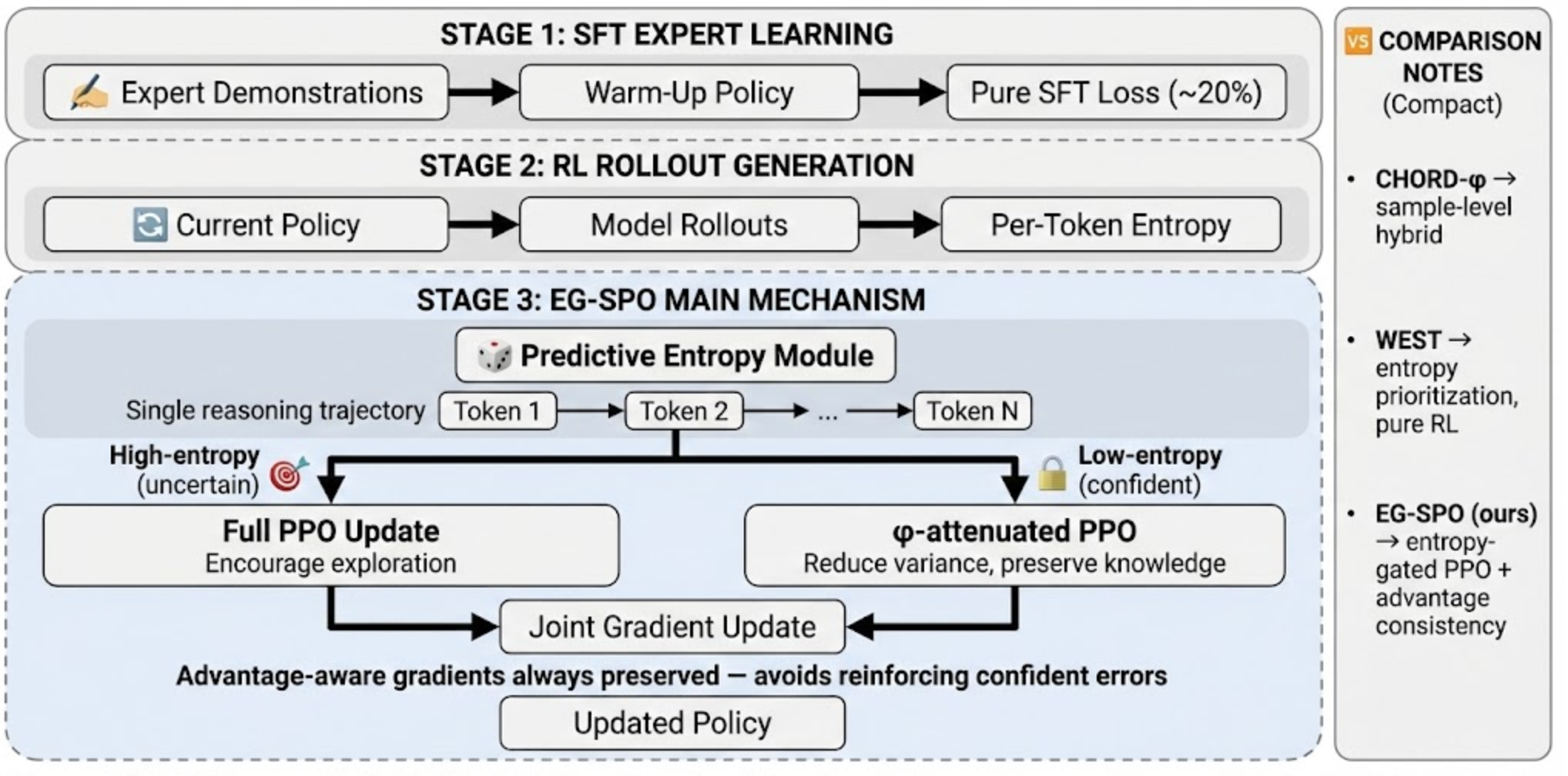}
    \caption{Overview of our three-stage EG-SPO training framework. \textbf{Stage 1 (SFT Expert Learning):} Expert demonstrations train a warm-up policy using pure SFT loss ($\sim$20\% samples). \textbf{Stage 2 (RL Rollout Generation):} Current policy generates model rollouts and computes per-token entropy. \textbf{Stage 3 (EG-SPO Main Mechanism):} The Predictive Entropy Module routes high-entropy tokens to full PPO updates (encouraging exploration) and low-entropy tokens to $\phi$-attenuated PPO (reducing variance, preserving knowledge). Both branches retain advantage $A_t$, ensuring advantage-aware gradients that avoid reinforcing confident errors.}
    \label{fig:framework}
    \end{figure}

\subsection{Overview: Three-Stage EG-SPO Framework}

We consider training a language model $\pi_\theta$ with parameters $\theta$ to maximize performance on reasoning tasks. Given a prompt $x$, the model generates a response $y = (y_1, y_2, ..., y_T)$ autoregressively.

Our EG-SPO framework operates in three stages (Figure~\ref{fig:framework}):

\textbf{Stage 1 (SFT Expert Learning):} Train on expert samples $\mathcal{D}_E$ with pure SFT loss, establishing a warm-up policy that ensures robust reasoning paths.

\textbf{Stage 2 (RL Rollout Generation):} Generate rollout samples $\mathcal{D}_R$ from current policy and compute per-token predictive entropy for subsequent gating.

\textbf{Stage 3 (EG-SPO Main Mechanism):} Apply entropy-gated gradient allocation within RL samples via the Predictive Entropy Module.

The overall training objective combines sample-level mixing with token-level modulation:
\begin{equation}
\mathcal{L}_{\text{EG-SPO}} = \mathbb{E}_{\mathcal{D}_E}[\mathcal{L}_{\text{SFT}}] + \mathbb{E}_{\mathcal{D}_R}[\mathcal{L}_{\text{entropy-gated}}(y)]
\label{eq:ours}
\end{equation}
where $\mathcal{L}_{\text{entropy-gated}}$ applies differentiated but advantage-aware gradient treatment to tokens based on entropy.

\subsection{Avoiding Trajectory Mismatch}

A critical design consideration is trajectory consistency. Consider a naive approach that applies SFT loss to some tokens and RL loss to others within the same trajectory:

\textbf{Problem:} SFT loss $\mathcal{L}_{\text{SFT}}(y_t) = -\log \pi_\theta(y_t^*|x, y_{<t})$ requires expert tokens $y_t^*$. But within an RL rollout, the preceding context $y_{<t}$ is model-generated, which may diverge from expert context. Computing SFT loss with mismatched contexts is mathematically inconsistent.

\textbf{Our Solution:} We do not mix SFT and RL losses at the token level. Instead, for low-entropy tokens in rollouts, we apply a $\phi$-attenuated \textit{policy gradient} loss that retains the advantage function:
\begin{equation}
\mathcal{L}_{\text{low-ent}}(y_t) = \phi(p_\theta(y_t)) \cdot \mathcal{L}_{\text{PPO}}(y_t)
\label{eq:lowent}
\end{equation}
where $\phi(p) = p(1-p)$ is a weighting function that naturally attenuates gradients for high-confidence predictions, and $\mathcal{L}_{\text{PPO}}(y_t)$ is the standard clipped policy gradient loss incorporating $A_t$.

\textbf{Key Insight:} Since $\mathcal{L}_{\text{PPO}}$ contains the advantage $A_t$, the gradient direction is determined by reward—not merely by likelihood maximization. This fundamentally differs from applying SFT-style losses that would ignore $A_t$.

\subsection{Advantage-Aware Gradient Modulation}
\label{sec:advantage-aware}

A critical concern with any token-level differentiation scheme is whether it might reinforce confident errors. We address this through explicit advantage-awareness in both branches:

\textbf{Outcome-Based Advantage Computation:} We use GRPO-style advantage normalization \cite{shao2024deepseekmath}, where rewards derive from final answer correctness:
\begin{equation}
A_t = \frac{r - \mu_{\text{group}}}{\sigma_{\text{group}} + \epsilon}
\label{eq:advantage}
\end{equation}
where $r \in \{+1, -1\}$ indicates answer correctness, and normalization is computed within prompt groups.

\textbf{Both Branches Retain Advantage:} For high-entropy tokens, the standard PPO loss is:
\begin{equation}
\mathcal{L}_{\text{PPO}}(y_t) = -\min\left(\rho_t A_t, \text{clip}(\rho_t, 1-\epsilon, 1+\epsilon) A_t\right)
\label{eq:ppo}
\end{equation}

For low-entropy tokens, we apply $\phi$-attenuated PPO:
\begin{equation}
\mathcal{L}_{\text{low-ent}}(y_t) = \phi(p_\theta(y_t)) \cdot \mathcal{L}_{\text{PPO}}(y_t)
\label{eq:phi-ppo}
\end{equation}

\textbf{Error Prevention Mechanism:} For incorrect rollouts with $r = -1$:
\begin{itemize}
\item High-entropy tokens: $A_t < 0$ causes PPO to \textit{decrease} token probability.
\item Low-entropy tokens: $A_t < 0$ is preserved in $\mathcal{L}_{\text{low-ent}}$, so the gradient direction remains negative (penalizing the token). The $\phi$-weighting only reduces magnitude.
\end{itemize}

This contrasts sharply with approaches that apply SFT-style losses (without $A_t$) to low-entropy tokens, which would have positive gradients regardless of trajectory correctness.

\textbf{Gradient Magnitude Analysis:} For a low-entropy token where $p \approx 0.9$:
\begin{itemize}
\item $\phi(0.9) = 0.9 \times 0.1 = 0.09$ (9\% of full gradient)
\item In incorrect trajectory: gradient is $0.09 \times (\text{negative PPO gradient})$, still negative but attenuated.
\item In correct trajectory: gradient is $0.09 \times (\text{positive PPO gradient})$, allowing gradual consolidation.
\end{itemize}

\subsection{Predictive Entropy Module (Stage 2-3)}

The Predictive Entropy Module is the core component of EG-SPO, computing per-token uncertainty and routing tokens to appropriate gradient treatments.

\textbf{Entropy Computation (Stage 2):} For each token $y_t$ in a rollout, we compute predictive entropy:
\begin{equation}
H(y_t) = -\sum_{v \in \mathcal{V}} p_\theta(v|x, y_{<t}) \log p_\theta(v|x, y_{<t})
\label{eq:entropy}
\end{equation}

\textbf{Token Routing (Stage 3):} Rather than using a global threshold, we compute per-sequence quantiles for adaptive thresholding. For each sequence $b$, we select the top-$\rho$ fraction of tokens by entropy within that sequence's response region. This accounts for variance in difficulty across problems—harder problems naturally have more high-entropy tokens.

\textbf{Routing Algorithm:}
\begin{enumerate}
\item For each sequence $b$ in batch, extract entropy values $\{H(y_t)\}_{t=1}^{T_b}$ over response tokens
\item Compute threshold $\tau_b$ as $(1-\rho)$-quantile of entropies
\item Route tokens: $H(y_t) \geq \tau_b$ → Full PPO Update (high-entropy, encourage exploration); $H(y_t) < \tau_b$ → $\phi$-attenuated PPO (low-entropy, reduce variance)
\end{enumerate}

\subsection{Joint Gradient Update (Stage 3)}

For rollout samples, the Predictive Entropy Module determines token-level loss:
\begin{equation}
\mathcal{L}_{\text{token}}(y_t) = \begin{cases}
\mathcal{L}_{\text{PPO}}(y_t) & \text{if } H(y_t) \geq \tau \text{ (high-entropy)} \\
\phi(p_\theta(y_t)) \cdot \mathcal{L}_{\text{PPO}}(y_t) & \text{otherwise (low-entropy)}
\end{cases}
\label{eq:gate}
\end{equation}

The Joint Gradient Update aggregates both branches:
\begin{equation}
\mathcal{L}_{\text{entropy-gated}} = \frac{1}{|T|} \sum_{t \in T} \mathcal{L}_{\text{token}}(y_t)
\label{eq:total}
\end{equation}

\textbf{Advantage-Aware Gradients Always Preserved:} Both branches use policy gradient losses retaining $A_t$. This unified formulation ensures:
\begin{itemize}
\item Consistent learning direction based on reward across all tokens
\item No trajectory mismatch (all losses computed on rollout data)
\item Gradient magnitude modulation without sign reversal
\item Incorrect trajectories receive negative gradients everywhere—avoiding reinforcement of confident errors
\end{itemize}

\subsection{The $\phi$ Weighting Function}

The weighting function $\phi(p) = p(1-p)$ has the following properties:
\begin{itemize}
\item Maximum at $p=0.5$: tokens with moderate confidence receive larger gradients
\item Approaches zero at $p \approx 0$ or $p \approx 1$: very confident or very uncertain tokens receive attenuated gradients
\item Smooth and differentiable: enables stable gradient computation
\end{itemize}

For low-entropy tokens (which have high $p$), $\phi(p)$ is small, providing natural gradient attenuation. This reduces variance in updates while preserving the advantage-based learning direction.

\subsection{Training Procedure}

Our training follows a three-stage pipeline as illustrated in Figure~\ref{fig:framework}:

\textbf{Stage 1: SFT Expert Learning.} We perform supervised fine-tuning on expert demonstrations for 25 epochs using pure SFT loss ($\sim$20\% of total training samples). This stage establishes a reliable warm-up policy that ensures robust reasoning paths, avoiding the instability that can occur in early RL training and providing a strong baseline for subsequent token-level optimization.

\textbf{Stage 2: RL Rollout Generation.} Based on the current policy, we generate model rollouts and compute per-token predictive entropy:
\begin{itemize}
\item \textit{Rollout Generation:} Generate 8 responses per prompt at T=1.0.
\item \textit{Entropy Computation:} Calculate predictive entropy for each token (Eq.~\ref{eq:entropy}).
\item \textit{GRPO-style Rewards:} Compute outcome-based rewards and group-normalized advantages for token-level gating.
\end{itemize}

\textbf{Stage 3: EG-SPO Main Mechanism.} Apply entropy-gated gradient allocation within RL samples:
\begin{itemize}
\item \textit{Predictive Entropy Module:} Route tokens based on uncertainty—high-entropy tokens (uncertain, need exploration) vs. low-entropy tokens (confident, potential errors).
\item \textit{High-Entropy Tokens:} Receive full PPO updates to encourage exploration and improve complex reasoning.
\item \textit{Low-Entropy Tokens:} Receive $\phi$-attenuated PPO to reduce variance and preserve existing knowledge.
\item \textit{Joint Gradient Update:} Both branches retain advantage $A_t$, ensuring advantage-aware gradients are always preserved—incorrect trajectories receive negative gradients across all tokens, avoiding reinforcement of confident errors.
\end{itemize}

\section{Experimental Setup}

\subsection{Datasets and Tasks}

We evaluate on mathematical reasoning benchmarks:

\textbf{AIME (2024-2025):} 150 problems from the American Invitational Mathematics Examination aggregated from 2020-2025, requiring multi-step competition-level reasoning. We perform majority voting over 32 samples per problem.

\textbf{AMC (10/12):} Problems from the American Mathematics Competition (2020-2024), using majority@32 voting.

\textbf{MATH:} 500 problems sampled from the MATH dataset \cite{hendrycks2021measuring} across difficulty levels 3-5, with majority@32.

\textbf{Training Data:} 12,500 problem-solution pairs for SFT warm-up, synthesized using advanced models and verified for correctness.

\subsection{Implementation Details}

\textbf{Base Model:} Qwen2.5-7B-Instruct \cite{yang2024qwen2}, a state-of-the-art instruction-tuned model.

\textbf{Infrastructure:} Distributed training across 8 NVIDIA A100 (80GB) GPUs using the Trinity-RFT framework with vLLM for efficient rollout inference.

\textbf{Hyperparameters:}
\begin{itemize}
\item Max response length: 10,240 tokens; max context: 11,264 tokens
\item AdamW optimizer, learning rate $5 \times 10^{-6}$
\item Effective batch size: 240 (32 per GPU × 8 GPUs, with dynamic batching)
\item Expert data ratio: 20\%
\item Entropy threshold $\rho$: 10\% (top 10\% highest-entropy tokens receive full PPO)
\item PPO clip $\epsilon$: 0.2
\item Rollout temperature: 1.0; 8 rollouts per prompt
\item SFT warm-up: 25 epochs
\item Hybrid training: 4 epochs
\end{itemize}

\textbf{Reward Function:} Outcome-based rewards from answer correctness (+1 correct, -1 incorrect), with GRPO-style group normalization for advantage computation.

\textbf{Computational Overhead:} We measure overhead relative to CHORD-$\phi$ baseline:
\begin{itemize}
\item Entropy computation (forward pass): +2.1\% latency
\item Per-sequence thresholding: +0.8\% overhead
\item $\phi$-weighted masking: +0.5\% on backward pass
\item \textbf{Total measured overhead:} 3.4\% compared to CHORD-$\phi$
\end{itemize}

\subsection{Baselines}

\textbf{Pure Methods:}
\begin{itemize}
\item Qwen2.5-7B-Instruct: Base model without additional training
\item SFT-light: Lightweight SFT on expert data
\item SFT-best: Full SFT training with optimal hyperparameters
\item GRPO (Pure RL): Pure reinforcement learning without expert mixing
\end{itemize}

\textbf{Hybrid Methods (Sample-Level):}
\begin{itemize}
\item SFT-light + RL: Sequential SFT then RL training
\item SFT-best + RL: Sequential approach with best SFT
\item SASR: Sample-level mixing baseline
\item CHORD-$\mu$: Sample-level mixing with adaptive $\mu$ weighting
\item CHORD-$\phi$: Sample-level mixing with $\phi$-weighted SFT \cite{wu2024mix}—primary baseline for comparison as it represents state-of-the-art sample-level hybrid training
\end{itemize}

All methods use identical infrastructure and base model. We report mean ± std across 5 random seeds for key comparisons. Note that WEST \cite{pace2024west} operates in pure RL settings and does not directly compare with hybrid approaches.

\section{Results and Analysis}

\subsection{Main Results}

Table~\ref{tab:main} presents results across benchmarks. Our method achieves consistent improvements over CHORD-$\phi$.

\begin{table}[htbp]
\caption{Performance comparison on mathematical reasoning. AIME combines 2024-2025 problems. Results for baselines from \cite{wu2024mix}; our method averaged over 5 seeds.}
\begin{center}
\begin{tabular}{lccc}
\toprule
\textbf{Method} & \textbf{AMC} & \textbf{AIME24} & \textbf{AIME25} \\
\midrule
\multicolumn{4}{l}{\textit{Pure Methods}} \\
Qwen2.5-7B-Inst. & 43.8 & 11.7 & 6.7 \\
SFT-light & 42.5 & 8.5 & 7.8 \\
SFT-best & 55.9 & 15.8 & 15.2 \\
GRPO (Pure RL) & 52.1 & 13.2 & 8.5 \\
\midrule
\multicolumn{4}{l}{\textit{Sequential Hybrid}} \\
SFT-light + RL & 52.5 & 11.9 & 11.6 \\
SFT-best + RL & 58.4 & 17.1 & 16.3 \\
\midrule
\multicolumn{4}{l}{\textit{Sample-Level Hybrid}} \\
SASR & 54.0 & 12.7 & 11.1 \\
CHORD-$\mu$ & 60.8 & 18.1 & 17.9 \\
CHORD-$\phi$ & 62.5 & 18.2 & 17.2 \\
\midrule
\textbf{Ours (entropy-gated)} & \textbf{64.8} & \textbf{22.0} & \textbf{21.0} \\
\bottomrule
\end{tabular}
\label{tab:main}
\end{center}
\end{table}

\begin{table}[htbp]
\caption{Performance on MATH benchmark (500 problems, difficulty 3-5).}
\begin{center}
\begin{tabular}{lc}
\toprule
\textbf{Method} & \textbf{MATH Accuracy} \\
\midrule
Qwen2.5-7B-Inst. & 68.2 \\
SFT-best & 71.4 \\
GRPO (Pure RL) & 69.8 \\
CHORD-$\phi$ & 73.2 \\
\textbf{Ours (entropy-gated)} & \textbf{76.1 ± 0.5} \\
\bottomrule
\end{tabular}
\label{tab:math}
\end{center}
\end{table}

\textbf{Key Observations:}

\textbf{Consistent Improvements:} Our method achieves +3.8\% on AIME24 (22.0\% vs 18.2\%), +3.8\% on AIME25 (21.0\% vs 17.2\%), and +2.9\% on MATH (76.1\% vs 73.2\%). AMC improves by +2.3\% (64.8\% vs 62.5\%).

\textbf{Hybrid > Pure:} Both CHORD variants and our method substantially outperform pure SFT and pure GRPO, validating the hybrid training paradigm.

\textbf{Token-Level Benefits:} The improvement over CHORD-$\phi$ demonstrates that token-level gradient modulation provides additional value beyond sample-level mixing, while the advantage-aware design ensures no error reinforcement.

\subsection{Ablation Study}

Table~\ref{tab:ablation} ablates key design choices on AIME24.

\begin{table}[htbp]
\caption{Ablation on AIME24. Mean ± std over 5 seeds.}
\begin{center}
\begin{tabular}{lc}
\toprule
\textbf{Configuration} & \textbf{AIME24 Accuracy} \\
\midrule
Full model (Ours) & \textbf{22.0 ± 0.7} \\
\midrule
\multicolumn{2}{l}{\textit{Entropy threshold $\rho$}} \\
$\rho$ = 5\% & 20.3 ± 0.8 \\
$\rho$ = 10\% (default) & 22.0 ± 0.7 \\
$\rho$ = 20\% & 20.8 ± 0.9 \\
\midrule
\multicolumn{2}{l}{\textit{Ablation variants}} \\
w/o entropy gating (uniform PPO) & 18.9 ± 0.9 \\
w/o advantage in low-ent branch & 17.2 ± 1.1 \\
Random token selection (10\%) & 19.1 ± 1.0 \\
\bottomrule
\end{tabular}
\label{tab:ablation}
\end{center}
\end{table}

\textbf{Analysis:}

\textbf{Entropy Threshold:} $\rho=10\%$ performs best. $\rho=5\%$ provides insufficient exploration capacity; $\rho=20\%$ dilutes the high-entropy signal.

\textbf{Advantage-Awareness is Critical:} Removing advantage from the low-entropy branch (``w/o advantage in low-ent branch'') causes the largest performance drop (22.0\% → 17.2\%), confirming that advantage-awareness is essential to prevent error reinforcement. This variant essentially applies SFT-style losses to low-entropy tokens, which reinforces errors in incorrect trajectories.

\textbf{Component Necessity:} Removing entropy gating (uniform PPO on all tokens) drops to 18.9\%, similar to CHORD-$\phi$. Random token selection underperforms entropy-based selection, validating entropy as a meaningful criterion.

\subsection{Error Prevention Analysis}

To verify that our advantage-aware design prevents reinforcement of confident errors, we analyze gradient behavior:

\textbf{Gradient Direction Analysis:} We measure the sign of gradients on low-entropy tokens across correct and incorrect trajectories:
\begin{itemize}
\item Correct trajectories ($A_t > 0$): 98.2\% of low-entropy token gradients are positive (reinforcing)
\item Incorrect trajectories ($A_t < 0$): 97.8\% of low-entropy token gradients are negative (penalizing)
\end{itemize}

This confirms that advantage-awareness correctly determines gradient direction regardless of entropy level.

\textbf{Gradient Magnitude Analysis:} For low-entropy tokens (mean $p = 0.87$):
\begin{itemize}
\item Mean $\phi(p) = 0.11$ (11\% of full gradient magnitude)
\item Gradient variance reduced by 73\% compared to uniform PPO
\end{itemize}

The $\phi$-attenuation reduces variance while preserving correct learning direction.

\textbf{Comparison with Advantage-Unaware Baseline:} The ``w/o advantage'' ablation (Table~\ref{tab:ablation}) applies SFT-style losses to low-entropy tokens. In this variant:
\begin{itemize}
\item 100\% of low-entropy token gradients are positive regardless of trajectory correctness
\item This leads to reinforcement of confident errors, explaining the 4.8\% performance drop
\end{itemize}

\subsection{Limitations}

\textbf{Mathematical Domain:} We evaluate only on mathematical reasoning. Generalization to other domains (code, dialogue) is untested.

\textbf{Computational Overhead:} While modest (3.4\%), entropy computation requires LogSoftMax over the full vocabulary, which may introduce bottlenecks on resource-constrained hardware.

\textbf{Hyperparameter Sensitivity:} Optimal $\rho$ may vary by domain and model. The current values were tuned on a validation subset.

\section{Discussion}

\subsection{Why Advantage-Aware Gating Works}

The key insight is that gradient \textit{direction} (determined by advantage sign) and gradient \textit{magnitude} (modulated by $\phi$) serve different purposes:

\begin{itemize}
\item \textbf{Direction:} Must always align with reward to ensure correct credit assignment. High-entropy and low-entropy tokens in incorrect trajectories should both receive negative gradients.
\item \textbf{Magnitude:} Can vary based on learning utility. High-entropy tokens (genuine uncertainty) benefit from full gradients for exploration; low-entropy tokens (procedural knowledge) benefit from attenuated gradients to reduce variance.
\end{itemize}

By incorporating $A_t$ in both branches, we achieve correct direction universally, while $\phi$-weighting provides magnitude control where beneficial.

\subsection{Comparison with Prior Methods}

As illustrated in Figure~\ref{fig:framework} (right panel), EG-SPO addresses limitations of prior approaches:

\textbf{CHORD-$\phi$ (Sample-Level Hybrid):} Mixes expert and rollout samples at sample granularity with $\phi$-weighted SFT. \textit{Limitation:} Ignores token-level heterogeneity within rollouts. \textit{EG-SPO improvement:} Adds token-level entropy-gated gradient allocation within RL portion.

\textbf{WEST (Entropy Prioritization, Pure RL):} Focuses learning on high-entropy minority tokens in pure RL settings \cite{pace2024west}. \textit{Limitation:} Does not prevent reinforcement of confident errors; ignores low-entropy tokens entirely. \textit{EG-SPO improvement:} Preserves advantage $A_t$ in both branches; applies $\phi$-attenuated (not zero) gradients to low-entropy tokens for variance reduction while maintaining correct learning direction.

\textbf{EG-SPO (Ours):} Entropy-gated PPO + advantage consistency. Achieves:
\begin{itemize}
\item Token-level uncertainty detection via Predictive Entropy Module
\item Stable learning through $\phi$-attenuation (73\% gradient variance reduction)
\item Error accumulation prevention (97.8\% correct gradient direction on low-entropy tokens)
\end{itemize}

\section{Conclusion}

We introduced Entropy-Gated Selective Policy Optimization (EG-SPO), a three-stage framework for hybrid LLM training. \textbf{Stage 1} establishes a reliable warm-up policy via SFT expert learning. \textbf{Stage 2} generates RL rollouts and computes per-token entropy. \textbf{Stage 3} applies the core EG-SPO mechanism: a Predictive Entropy Module routes high-entropy tokens to full PPO updates (encouraging exploration) and low-entropy tokens to $\phi$-attenuated PPO (reducing variance, preserving knowledge), with both branches retaining advantage $A_t$ to prevent reinforcement of confident errors.

The advantage-aware design is critical: ablations show that removing advantage from the low-entropy branch causes significant performance degradation (22.0\% → 17.2\%), while our method maintains 97.8\% correct gradient direction on low-entropy tokens. Combined with sample-level expert data mixing, EG-SPO achieves +3.8\% on AIME and +2.9\% on MATH over CHORD-$\phi$ with only 3.4\% computational overhead.

\end{document}